\documentclass{article}
\usepackage{spconf,amsmath,graphicx}
\usepackage{amssymb,amsfonts}
\usepackage{xcolor}
\usepackage{array}
\usepackage{threeparttable}
\usepackage{hyperref}
\usepackage{multirow}

\newcolumntype{P}[1]{>{\centering\arraybackslash}p{#1}}
\newcolumntype{L}[1]{>{\arraybackslash}p{#1}}

\title{FD-MobileNet: Improved MobileNet with a Fast Downsampling Strategy}

\name{Zheng Qin, Zhaoning Zhang, Xiaotao Chen, Yuxing Peng\thanks{Correspondence should be addressed to Zhaoning Zhang (Email: \url{zzningxp@gmail.com}).}}
\address{Science and Technology on Parallel and Distributed Laboratory,\\
National University of Defense Technology, Changsha, China}

\begin{document}
\ninept

\hyphenation{networks methods para-meters appears op-tical net-works semi-conduc-tor store stor-age snap-shot space reads among Ursa exits scal-able uses sys-tems IOPS dif-ferent GFS HDFS FS gen-erally SCSI NFS HBase SSD slightly stored two anony-mized speak-ing lineariza-bility server elim-inate states pro-cesses signifi-cantly sin-gle OCFS blocks par-allelism parallel-ism returns pre-senting present-ing stand-ard stores IOPS servers appends scala-bility pri-mary mode respec-tively re-spectively sat-isfy UPSs CPUs avail-ability availa-bility availabil-ity clouds usu-ally against re-quests inde-pendently namely storing Meituan evolv-ing since ob-ject tests BS mounted under-loaded consis-tency status jour-nals place-ment statis-tics exclu-sively meta-data mainly collabora-tively collabo-ratively FDS focus nodes remote conven-tional strata single enough daily require small sequen-tial Theore-tically coroutine acce-leration accele-ration effi-cient matrix vectors imple-mentations implemen-tations implementa-tions larger Mobile-Nets GPUs com-puted cuDNN hyper-parameters hyperpara-meters Con-sequently Conse-quently app-lications appli-cations applica-tions adopted GPU accele-rating acce-lerating Tensor-Flow results frame-works convo-lutions con-volutions convolu-tions convo-lution con-volution convolu-tion threads signi-ficant signifi-cant sig-nificant ii iii GEMM aca-demic acade-mic its weights algo-rithms uti-lize also rela-tively redun-dancy imple-mentation implemen-tation implementa-tion Mobile-Net multi-plier fewer res-pectively respec-tively kernel table uti-lizes every in-creases diag-onalwise Multiplications be-nefit be-nefits mo-dification suf-fer in-duced shu-ffle evo-lution mo-dules alle-viating discrimi-native com-putational ope-rations mo-dule fa-mily fami-ly fur-ther ope-ration chan-nels infor-mative in-formative capa-city capaci-ty memo-ry me-mory capa-bility capabi-lity bypass stra-tegy convo-lutions}

\hyphenpenalty=1000

\maketitle

\begin{abstract}
We present \emph{Fast-Downsampling MobileNet} (\emph{FD-MobileNet}), an efficient and accurate network for very limited computational budgets (e.g., 10-140 MFLOPs).
Our key idea is applying a fast downsampling strategy to MobileNet framework.
In FD-MobileNet, we perform 32$\times$ downsampling within 12 layers, only half the layers in the original MobileNet.
This design brings three advantages:
(i) It remarkably reduces the computational cost.
(ii) It increases the information capacity and achieves significant performance improvements.
(iii) It is engineering-friendly and provides fast actual inference speed.
Experiments on ILSVRC 2012 and PASCAL VOC 2007 datasets demonstrate that FD-MobileNet consistently outperforms MobileNet and achieves comparable results with ShuffleNet under different computational budgets, for instance, surpassing MobileNet by 5.5\% on the ILSVRC 2012 top-1 accuracy and 3.6\% on the VOC 2007 mAP under a complexity of 12 MFLOPs.
On an ARM-based device, FD-MobileNet achieves 1.11$\times$ inference speedup over MobileNet and 1.82$\times$ over ShuffleNet under the same complexity.
\end{abstract}


\section{Introduction}

Deep convolutional neural networks (CNNs) have become one of the most important methods in computer vision tasks such as image classification \cite{krizhevsky2012imagenet,simonyan2014very,he2016deep,huang2016densely}, object detection \cite{ren2015faster,liu2016ssd,redmon2016yolo9000,he2017mask} and semantic segmentation \cite{long2015fully,chen2016deeplab}.
However, state-of-the-art CNNs require enormous computational resources and huge model sizes, which prevents them from being deployed on mobile or embedded devices.

For this reason, the inference-time compression and acceleration of deep neural networks has attracted the attention of the deep learning community in recent years.
The related work is conventionally categorized into four classes.
\emph{Tensor decomposition} methods \cite{jaderberg2014speeding,zhang2016accelerating} factorize a convolutional layer into several smaller convolutional layers, which reduces the overall complexity and the number of parameters.
This class of methods conventionally involve a low-rank estimation process and a fine-tuning process, leading to a slow training procedure.
\emph{Parameter quantization} methods \cite{vanhoucke2011improving,courbariaux2016binarized} propose to utilize low-precision parameters in neural networks and provide significant theoretical speedup and enormous memory savings.
However, current hardware is not well optimized for low-precision computation so specific hardware is required for quantization methods to achieve an ideal speedup.
\emph{Network pruning} methods \cite{han2015learning,liu2017learning} attempt to discover and alleviate parameter and structure redundancy in deep neural networks.
Early pruning approaches adopt an unstructured pruning scheme and induces random memory accesses, which is not well supported by current hardware.
Recent research on network pruning mainly focuses on structured pruning to leverage existing hardware.
At last, \emph{compact networks} \cite{iandola2016squeezenet,howard2017mobilenets,zhang2017shufflenet} are specifically designed to employ both accurate and computationally economical networks on mobile or embedded devices.

Unlike the other methods which are mainly focused on compressing pre-trained models, compact networks can be trained from scratch.
Additionally, compact networks are orthogonal to the other methods and can be further accelerated.
In view of these advantages, various compact network architectures have been proposed.
Among these networks, MobileNet \cite{howard2017mobilenets} and ShuffleNet \cite{zhang2017shufflenet} achieve the state-of-the-art performance.

ShuffleNet is composed of a variant of the bottleneck unit \cite{he2016deep} named the ShuffleNet unit.
The ShuffleNet unit utilizes bypass connections for better representation capability.
Beneficial from the powerful ShuffleNet unit, ShuffleNet achieves significant performance improvements over previous architectures \cite{iandola2016squeezenet,howard2017mobilenets}.
However, the bypass connection structure introduces multiple information paths in the computing graph, which induces frequent memory/cache switches in the engineering implementation on mobile or embedded devices.
Consequently, the actual inference speed of ShuffleNet on physical devices is not ideal.

On the contrary, MobileNet exploits depthwise separable convolutions as its building blocks in a simple stacking architecture.
This design allows a more efficient utilization of memory and cache, and MobileNet is significantly faster than ShuffleNet in actual inference speed under the same complexity.
However, MobileNet adopts a slow downsampling strategy, which induces severe performance degradation when the computational budget is relatively small, for instance, 10-140 MFLOPs.
In such a slow downsampling strategy, more layers have large feature maps, so the feature representation is more detailed.
However, the number of channels in the network is restricted, thus the information capacity is relatively small.
If the width of a network is further shrunk to fit an extremely limited complexity, the information capacity will become too small and the performance of the network will collapse.


In this paper, we present a highly efficient and accurate network named \emph{Fast-Downsampling MobileNet} (\emph{FD-MobileNet}) for extremely limited computational resources (e.g., 10 to 140 MFLOPs).
Instead of merely shrinking the width of the network to fit small computational budgets, we compose FD-MobileNet by adopting a fast downsampling strategy into the MobileNet framework.
In the proposed FD-MobileNet, we perform 32$\times$ downsampling within the first 12 layers, which is only half of the number in the original MobileNet.
After that, a sequence of depthwise separable convolutions are applied for better representation capability.
Benefiting from the fast downsampling strategy, FD-MobileNet has the following three advantages:
(i) The computational cost of FD-MobileNet is reduced as the spatial dimensions of the feature maps are smaller.
(ii) FD-MobileNet allows more channels than the MobileNet counterpart under the same complexity.
This remarkably increases the information capacity of FD-MobileNet, which is critical to the performance of very small networks.
(iii) FD-MobileNet inherits the simple architecture from MobileNet and provides a fast inference speed in engineering implementation.

We conduct extensive experiments to examine the effectiveness of the proposed FD-MobileNet.
Firstly, we compare FD-MobileNet with other state-of-the-art compact networks on the ILSVRC 2012 dataset \cite{russakovsky2015imagenet}.
Then, we examine the generalization ability of FD-MobileNet on the PASCAL VOC 2007 dataset \cite{everingham2010pascal}.
Experiments show that the proposed FD-MobileNet significantly outperforms MobileNet and achieves comparable performance with ShuffleNet under various computational budgets.
For instance, FD-MobileNet achieves improvements of 5.5\% on the ILSVRC 2012 top-1 accuracy and 3.6\% on the VOC 2007 mAP over MobileNet under the computational budget of 12 MFLOPs.
At last, we furthermore evaluate the actual inference speed of FD-MobileNet on an ARM-based device.
Under a complexity of 12 MFLOPs, FD-MobileNet provides 1.11$\times$ speedup over MobileNet and 1.82$\times$ over ShuffleNet.
Our code will be made publicly available later.

\section{Fast-Downsampling MobileNet}

In this section, we present the design of \emph{Fast-Downsampling MobileNet} (\emph{FD-MobileNet}).
FD-MobileNet is composed of the highly efficient depthwise separable convolutions and adopts a fast downsampling strategy.
Benefiting from this design, FD-MobileNet achieves both high accuracy and high efficiency under very limited computational budgets.

\noindent\textbf{Depthwise Separable Convolutions.}
Following MobileNet \cite{howard2017mobilenets}, FD-MobileNet exploits depthwise separable convolutions \cite{chollet2016xception} as the building blocks.
A $k \times k$ depthwise separable convolution factorizes a $k \times k$ standard convolution into a $k \times k$ depthwise convolution and a pointwise convolution with 8$\sim$9 times reduction in FLOPs.
In practice, depthwise separable convolutions can achieve comparable performance with standard convolutions while provide great efficiency on computation-limited devices.

\begin{figure}[!t]
\centering
\includegraphics[width=0.4\textwidth]{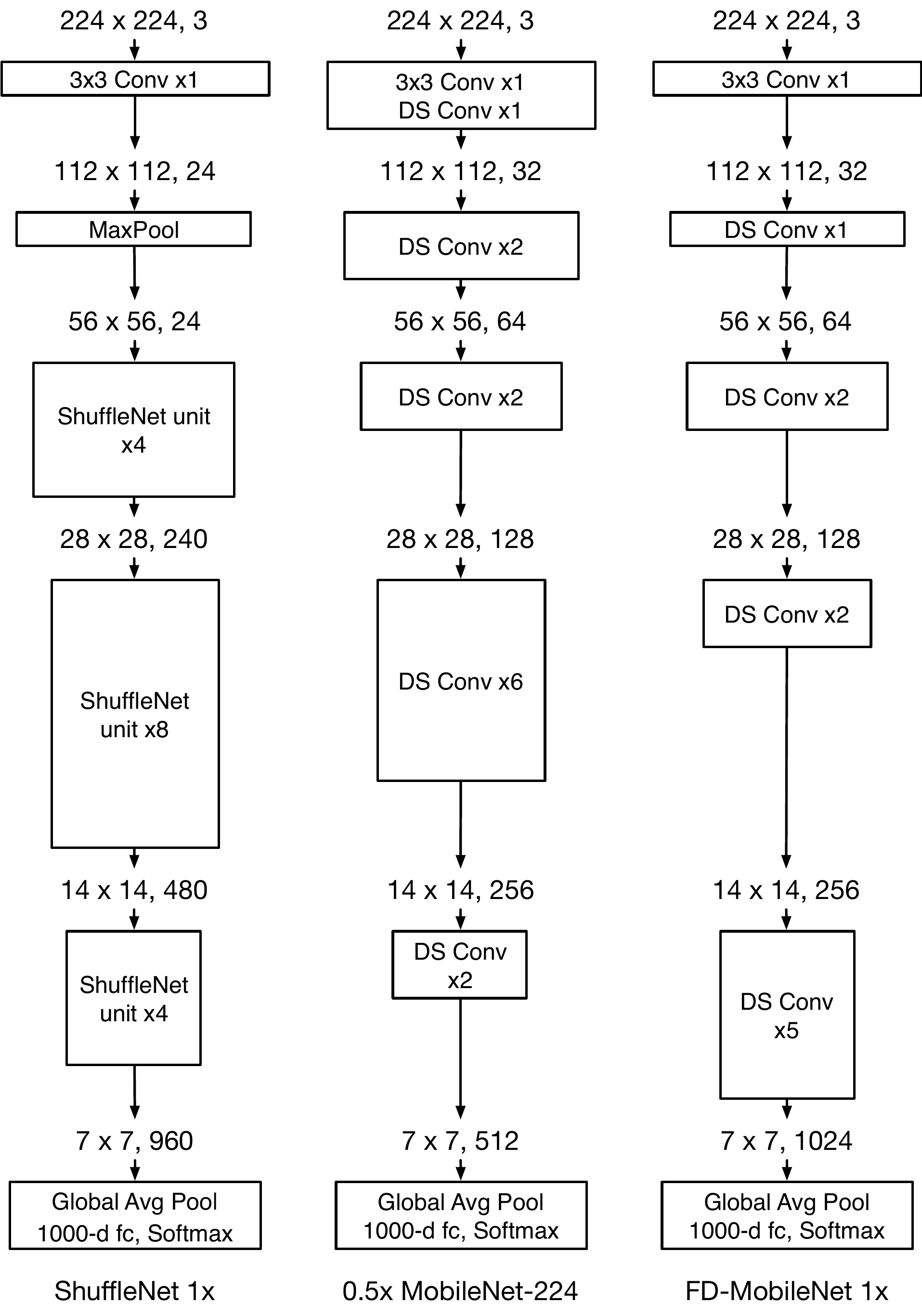}
\caption{Comparison of the downsampling strategies of FD-MobileNet, MobileNet and ShuffleNet under a complexity of 140 MFLOPs. The width of each block (except the ones at the bottom) represents the spatial dimensions while the height represents the number of building blocks. Compared with MobileNet and ShuffleNet, FD-MobileNet adopts a much faster downsampling strategy and leverages more channels, which enlarges the information capacity and gains performance improvements. \textbf{DS Conv}: depthwise separable convolution. Each depthwise separable convolution consists of two layers while each ShuffleNet unit has three layers.}
\label{figure:architecture}
\end{figure}

\noindent\textbf{Fast Downsampling Strategy.}
Modern CNNs adopt a hierarchical architecture, where the spatial dimensions of the layers within the same stage is kept identical, and the spatial dimensions in the next stage is reduced by downsampling.
In view of the restricted computational budgets, compact networks suffer from both the weak feature representation and the restricted information capacity.
Different downsampling strategies provide a trade-off between detailed feature representation and large information capacity for compact networks.
In a \emph{slow downsampling strategy}, downsampling is performed in the later layers of the network, thus more layers have large spatial dimensions.
On the contrary, downsampling is performed at the beginning of the network in a \emph{fast downsampling strategy}, which significantly reduces the computational cost.
Consequently, given a fixed computational budget, a slow downsampling strategy is inclined to generate more detailed features, whereas a fast downsampling strategy can increase the number of channels and allows more information to be encoded.

When the computational budget is extremely small, the information capacity plays a more important role in the performance of a network.
Conventionally, the number of channels is reduced to adapt a compact network architecture to a certain complexity.
In the case where a slow downsampling scheme is adopted, the network becomes too narrow to encode adequate information, which induces severe performance degradation.
For instance, under a complexity of 12 MFLOPs, the original MobileNet architecture only has 128 channels in the last layer before the global pooling, thus the information capacity is very limited.

Based on this insight, we propose to adopt a fast downsampling strategy in the architecture of FD-MobileNet and postpone the feature extraction process to the smallest resolution.
The faster downsampling is implemented by consecutively applying depthwise separable convolutions with large strides at the beginning of the network.
Here we do not use max pooling because we find it does not gain performance improvements but introduces extra computation.
The proposed FD-MobileNet accepts an image with a size of 224$\times$224 pixels, and performs 4$\times$ downsampling within the first 2 layers while performs 32$\times$ downsampling within merely 12 layers, whereas the number of layers performing the same downsampling in the original MobileNet is 4 and 24, respectively.
More specifically, the 12 layers are composed of 1 standard convolutional layer, 5 depthwise separable convolutions (each has a depthwise convolutional layer and a pointwise convolutional layer), and 1 depthwise convolutional layer.
Fig.~\ref{figure:architecture} illustrates the comparison of the downsampling strategies of FD-MobileNet, MobileNet and ShuffleNet under the computational budget of 140 MFLOPs.
From the figure, it is observed that FD-MobileNet is significantly shallower than the other architecture before the feature maps are shrunk to 7$\times$7.


\begin{table}[!t]
\centering
\caption{Fast-Downsampling MobileNet architecture. ``\textbf{/2}'' indicates the stride of the layer is 2. \textbf{DWConv}: depthwise convolution.}
\label{table:architecture}
\begin{tabular}{|c|l|c|}
\hline
\textbf{Output Size} & \textbf{Layer} & \textbf{MFLOPs} \\ \hline \hline
$224\times224$ & Image & \\ \hline
$112\times112$ & $3\times3$ Conv, 32, /2 & 10.8 \\ \hline
\multirow{2}{*}{$56\times56$} & $3\times3$ DWConv, 32, /2 & \multirow{2}{*}{7.3} \\ \cline{2-2}
 & $1\times1$ Conv, 64 & \\ \hline
\multirow{4}{*}{$28\times28$} & $3\times3$ DWConv, 64, /2 & \multirow{4}{*}{20.6} \\ \cline{2-2}
 & $1\times1$ Conv, 128 & \\ \cline{2-2}
 & $3\times3$ DWConv, 128 & \\ \cline{2-2}
 & $1\times1$ Conv, 128 & \\ \hline
\multirow{4}{*}{$14\times14$} & $3\times3$ DWConv, 128, /2 & \multirow{4}{*}{19.9} \\ \cline{2-2}
 & $1\times1$ Conv, 256 & \\ \cline{2-2}
 & $3\times3$ DWConv, 256 & \\ \cline{2-2}
 & $1\times1$ Conv, 256 & \\ \hline
\multirow{6}{*}{$7\times7$} & $3\times3$ DWConv, 256, /2 & \multirow{6}{*}{84.7} \\ \cline{2-2}
 & $1\times1$ Conv, 512 & \\ \cline{2-2}
 & $4\times \begin{array}{l} 3\times3 \text{ DWConv, 512} \\ 1\times1 \text{ Conv, 512} \end{array}$ & \\ \cline{2-2}
 & $3\times3$ DWConv, 512 & \\ \cline{2-2}
 & $1\times1$ Conv, 1024 & \\ \hline
\multirow{2}{*}{$1\times1$} & Global Average Pooling & \multirow{2}{*}{1.0} \\ \cline{2-2}
 & 1000-d fc, Softmax & \\ \hline
\end{tabular}
\end{table}

\noindent\textbf{Remaining Layers.}
The utilization of the fast downsampling strategy significantly reduces the computation cost of the layers before the smallest spatial dimensions (7$\times$7).
Under the computational budget of 140 MFLOPs, MobileNet spends about 129 MFLOPs on the largest 4 resolutions, whereas FD-MobileNet only spends about 59 MFLOPs, as shown in Table~\ref{table:architecture}.
Consequently, more layers and more channels can be leveraged in the proposed architecture.
Here we exploit 6 depthwise separable convolutions to improve the representation capability of generated features.
The output channels of the first 5 depthwise separable convolutions are 512, while the last one is 1024, which is twice the number in the MobileNet counterpart (0.5$\times$ MobileNet-224).
The increase in the number of channels contributes to larger information capacity, which is critical to the performance of the networks under extremely limited computational resources.


\noindent\textbf{Overall Architecture.}
The overall architecture of FD-MobileNet is demonstrated in Table~\ref{table:architecture}.
FD-MobileNet adopts a simple stacking architecture with 24 layers, including 1 standard convolutional layer, 11 depthwise separable convolutions, and 1 fully-connected layer.
Following \cite{howard2017mobilenets}, a batch normalization \cite{ioffe2015batch} and a ReLU activation is applied after each convolutional layer.
To conveniently adapt FD-MobileNet to different computational budgets, we introduce a hyper-parameter $\alpha$ termed \emph{width multiplier} as in \cite{howard2017mobilenets} to uniformly adjust the width of FD-MobileNet.
We use a simple notation ``FD-MobileNet $\alpha\times$'' to represent a network with a width multiplier $\alpha$, and the network in Table~\ref{table:architecture} is denoted as ``FD-MobileNet 1$\times$''.

\noindent\textbf{Inference Efficiency.}
Current deep learning frameworks accomplish the inference of a neural network by building an acyclic computing graph.
For mobile or embedded devices, memory and cache resources are limited.
As a result, complicated computing graphs can induce frequent memory/cache switches, which slows down the actual inference speed.
FD-MobileNet inherits the simple architecture of the original MobileNet, and there is only one information path in the computing graph.
This makes FD-MobileNet very friendly to engineering implementation and efficient on physical devices.

\section{Experiments}


\subsection{Results on ILSVRC 2012 dataset}

We first evaluate the effectiveness of FD-MobileNet on the ILSVRC 2012 dataset \cite{russakovsky2015imagenet}.
The ILSVRC 2012 dataset is composed of 1.2 million training images and 50,000 validation images.
In the experiments, the networks are trained on the training set using PyTorch \cite{collobert2011torch7} with four GPUs for 90 epochs.
Following \cite{he2016deep}, the batch size is set to 256 and a momentum of 0.9 is used.
The learning rate starts from 0.1 and decays by an order of magnitude every 30 epochs.
As the networks are relatively small, a weight decay of 4e-5 is utilized as recommended in \cite{zhang2017shufflenet}.
For data augmentation, we adopt a slightly less aggressive multi-scale augmentation scheme without using color jittering.
On evaluation, the center-crop top-1 accuracy rates on the validation set are reported.
Each validation image is first resized with its shorter edge to 256 pixels, and then evaluated using the center $224 \times 224$ pixels crop.
Table~\ref{table:ilsvrc-2012} demonstrated the comparison of the top-1 accuracy of FD-MobileNet, MobileNet and ShuffleNet under three computational budgets.

\begin{table}[!t]
\centering\caption{Top-1 Accuracy (\%, \emph{larger is better}) on ILSVRC 2012 dataset. We re-implement MobileNet under a complexity of 12 MFLOPs as no results are reported in \cite{howard2017mobilenets}}
\label{table:ilsvrc-2012}
\begin{tabular}{|l|P{4em}|c|}
\hline
\textbf{Models} & \textbf{MFLOPs} & \textbf{Top-1 Acc.} \\ \hline \hline
ShuffleNet 1$\times$ \cite{zhang2017shufflenet} & 137 & \textbf{65.9} \\
0.5$\times$ MobileNet-224 \cite{howard2017mobilenets} & 149 & 63.7 \\
FD-MobileNet 1$\times$ (\textit{ours}) & 144 & \textbf{65.3} \\ \hline
ShuffleNet 0.5$\times$ \cite{zhang2017shufflenet} & 38 & \textbf{57.3} \\
0.25$\times$ MobileNet-224 \cite{howard2017mobilenets} & 41 & 50.6 \\
FD-MobileNet 0.5$\times$ (\textit{ours}) & 40 & \textbf{56.2} \\ \hline
ShuffleNet 0.25$\times$ \cite{zhang2017shufflenet} & 13 & \textbf{46.7} \\
0.125$\times$ MobileNet-224 \cite{howard2017mobilenets} & 12 & 39.6 \\
FD-MobileNet 0.25$\times$ (\textit{ours}) & 12 & \textbf{45.1} \\ \hline
\end{tabular}
\end{table}

\begin{table*}[!t]
\centering
\scriptsize
\setlength{\tabcolsep}{2.5pt}
\caption{mAP (\%, \emph{larger is better}) and AP (\%, \emph{larger is better}) on PASCAL VOC 2007 test set (600$\times$ resolution).}
\label{table:voc-2007}
\begin{tabular}{|l|c|c|c|c|c|c|c|c|c|c|c|c|c|c|c|c|c|c|c|c|c|}
\hline
\textbf{Backbone} & \textbf{mAP} & \textbf{areo} & \textbf{bike} & \textbf{bird} & \textbf{boat} & \textbf{bottle} & \textbf{bus} & \textbf{car} & \textbf{cat} & \textbf{chair} & \textbf{cow} & \textbf{table} & \textbf{dog} & \textbf{horse} & \textbf{mbike} & \textbf{person} & \textbf{plant} & \textbf{sheep} & \textbf{sofa} & \textbf{train} & \textbf{tv} \\ \hline \hline
0.5$\times$ MobileNet-224 \cite{howard2017mobilenets} & 53.8 & \textbf{59.0} & 66.8 & \textbf{52.3} & \textbf{33.5} & \textbf{29.8} & 56.9 & \textbf{71.4} & 61.5 & 29.8 & 59.2 & 51.3 & 59.3 & 69.9 & 64.6 & 63.5 & \textbf{29.5} & 48.9 & 51.8 & \textbf{65.0} & 52.1 \\
FD-MobileNet 1$\times$ (\textit{ours}) & \textbf{55.4} & 58.1 & \textbf{67.1} & 49.4 & 32.7 & 28.8 & \textbf{62.2} & 71.1 & \textbf{67.2} & \textbf{32.6} & \textbf{59.4} & \textbf{58.0} & \textbf{63.0} & \textbf{72.3} & \textbf{65.7} & \textbf{65.8} & 26.9 & \textbf{53.5} & \textbf{51.9} & \textbf{65.0} & \textbf{56.7} \\ \hline
0.25$\times$ MobileNet-224 \cite{howard2017mobilenets} & 42.3 & \textbf{47.5} & \textbf{53.8} & 35.0 & 24.0 & \textbf{18.5} & 43.9 & 60.2 & 51.3 & 17.5 & 47.6 & 47.5 & \textbf{47.4} & 60.0 & 58.7 & 55.5 & 19.2 & 38.3 & 36.3 & 48.9 & 34.2 \\
FD-MobileNet 0.5$\times$ (\textit{ours}) & \textbf{45.1} & 46.4 & 53.2 & \textbf{38.2} & \textbf{29.3} & 16.8 & \textbf{47.1} & \textbf{63.0} & \textbf{56.2} & \textbf{22.3} & \textbf{48.8} & \textbf{49.4} & 47.3 & \textbf{66.9} & \textbf{60.6} & \textbf{56.8} & \textbf{20.0} & \textbf{44.5} & \textbf{40.9} & \textbf{57.0} & \textbf{38.0} \\ \hline
0.125$\times$ MobileNet-224 \cite{howard2017mobilenets} & 29.1 & 33.4 & 38.6 & 20.7 & 16.0 & 2.9 & 31.4 & 48.5 & \textbf{42.7} & 13.2 & 26.8 & 28.2 & 34.5 & 46.9 & 45.4 & \textbf{42.3} & \textbf{13.4} & 29.3 & 21.7 & 29.6 & 16.1 \\
FD-MobileNet 0.25$\times$ (\textit{ours}) & \textbf{32.7} & \textbf{40.6} & \textbf{43.6} & \textbf{21.4} & \textbf{16.2} & \textbf{8.2} & \textbf{33.7} & \textbf{50.7} & 41.2 & \textbf{15.6} & \textbf{37.2} & \textbf{33.4} & \textbf{36.2} & \textbf{54.7} & \textbf{50.4} & 41.4 & 8.1 & \textbf{29.6} & \textbf{25.3} & \textbf{47.7} & \textbf{18.2} \\ \hline
\end{tabular}
\end{table*}

From the table, FD-MobileNet achieves substantial improvements over MobileNet under different computational budgets.
It is observed that FD-MobileNet surpasses MobileNet by a margin of 1.6\% under a complexity of 140 MFLOPs, and performs 5.6\% and 5.5\% better when the computational budget is 40 and 12 MFLOPs, respectively.
It is noteworthy that FD-MobileNet provides significantly improvements over MobileNet when the computational budget is very small (e.g., 40 and 12 MFOPs).
We attribute these improvements to the effectiveness of the fast downsampling strategy in FD-MobileNet.
The original MobileNet adopts a slow downsampling strategy, thus more layers have relatively large feature maps and are more computationally intensive.
Consequently, MobileNet is relatively narrow to maintain computational efficiency, which limits the information capacity.
On the other side, FD-MobileNet exploits a much faster downsampling strategy, which allows more channels to be leveraged and alleviates the information capacity degradation.
For instance, under 12 MFLOPs, the last layer of MobileNet outputs only 128 channels, whereas the number in FD-MobileNet is doubled.
The increase in the information capacity significantly improves the performance of FD-MobileNet.

Compared with ShuffleNet, FD-MobileNet achieves comparable or slightly worse results.
We conjecture that these differences are owed to the effectiveness of the bypass connection structure of the ShuffleNet unit.
The bypass connection structure has proven powerful in various computer vision tasks \cite{he2016deep,ren2015faster,he2017mask}.
However, on low-power mobile or embedded devices, the bypass connection structure induces frequent memory/cache switches and harms the actual inference speed.
On the contrary, the simple architecture of FD-MobileNet contributes to an efficient utilization of memory and cache.
Details are discussed in Section~\ref{section:actual-inference-time}.

\subsection{Results on PASCAL VOC 2007 dataset}
\label{section:voc-2007}


We furthermore conduct extensive experiments on PASCAL VOC 2007 detection dataset \cite{everingham2010pascal} to examine the generalization ability of the proposed FD-MobileNet.
PASCAL VOC 2007 dataset consists of about 10,000 images split into three (train/val/test) sets.
In the experiments, the detectors are trained on VOC 2007 trainval set, and the single-model results on VOC 2007 test set are reported.
We adopt the Faster R-CNN detection pipeline \cite{ren2015faster} and compare the performance of FD-MobileNet and MobileNet on 600$\times$ resolution under three computational budgets (140, 40 and 12 MFLOPs).
The detectors are trained for 15 epochs with a batch size of 1.
The learning rate starts from 1e-3, and is divided by 10 every 5 epochs.
The weight decay is set to 4e-5.
Other hyper-parameter settings follow the original Faster R-CNN in \cite{ren2015faster}.
During testing, 300 proposals are sent to the R-CNN subnet to generate the final predictions.

The comparison of the results are demonstrated in Table~\ref{table:voc-2007}.
It is observed that FD-MobileNet achieves significant improvements over MobileNet under different computational budgets.
Under the computational budget of 140 MFLOPs, the FD-MobileNet detector surpasses the MobileNet detector by a margin of 1.6\% on mAP.
The gap is enlarged when the complexity is lower.
When the complexity is restricted to 40 and 12 MFLOPs, FD-MobileNet outperforms MobileNet by 2.8\% and 3.6\% on mAP, respectively.
More specifically, on single class results, FD-MobileNet performs better than MobileNet on most classes.
From Table~\ref{table:voc-2007}, FD-MobileNet provides more significant improvements over MobileNet when the computational budget is smaller.
For instance, when the computational budget is 12 MFLOPs, FD-MobileNet achieves consistent improvements on the classes which are hard for MobileNet, such as bottle (5.3\%), chair (2.4\%) and boat (0.2\%).
These improvements have proven that FD-MobileNet have strong generalization ability for transfer learning.

\subsection{Actual Inference Time Evaluation}
\label{section:actual-inference-time}

\begin{table}[!t]
\centering
\caption{Actual inference time (ms, \emph{smaller is better}) on an ARM-based device with NCNN.}
\label{table:actual-inference-time}
\begin{tabular}{|l|P{4em}|c|}
\hline
\textbf{Models} & \textbf{MFLOPs} & \textbf{Time} \\ \hline \hline
ShuffleNet 1$\times$ \cite{zhang2017shufflenet} & 137 & 522.27 \\
0.5$\times$ MobileNet-224 \cite{howard2017mobilenets} & 149 & 431.73 \\
FD-MobileNet 1$\times$ (\textit{ours}) & 144 & \textbf{391.66} \\ \hline
ShuffleNet 0.5$\times$ \cite{zhang2017shufflenet} & 38 & 204.97 \\
0.25$\times$ MobileNet-224 \cite{howard2017mobilenets} & 41 & 155.84 \\
FD-MobileNet 0.5$\times$ (\textit{ours}) & 40 & \textbf{139.47} \\ \hline
ShuffleNet 0.25$\times$ \cite{zhang2017shufflenet} & 13 & 103.79 \\
0.125$\times$ MobileNet-224\tnote{*} \cite{howard2017mobilenets} & 12 & 63.73 \\
FD-MobileNet 0.25$\times$ (\textit{ours}) & 12 & \textbf{57.17} \\ \hline
\end{tabular}
\end{table}

To investigate the performance on physical devices, we further compare the actual inference time of FD-MobileNet, MobileNet and ShuffleNet on an ARM-based platform.
The experiments are conducted using an optimized NCNN framework \cite{ncnn} on an i.MX 6 series CPU (single-core, 800 MHz).

Table~\ref{table:actual-inference-time} shows the inference time of the three compact networks under computational budgets of 140, 40 and 12 MFLOPs, respectively.
Compared with MobileNet, FD-MobileNet achieves about 1.1$\times$ speedup over MobileNet under the three computational budgets.
These improvements are attributed to the effectiveness of the fast downsampling architecture of FD-MobileNet.
Compared with ShuffleNet, FD-MobileNet provides significantly faster inference speed.
When the computational budgets are 140 and 40 MFLOPs, FD-MobileNet gains 1.33$\times$ and 1.47$\times$ speedup over ShuffleNet, respectively.
The speedup is elevated under a complexity of 12 MFLOPs: FD-MobileNet is 1.82$\times$ faster than ShuffleNet.
It is noteworthy that under 140 and 40 MFLOPs, the ShuffleNet models have fewer FLOPs than the FD-MobileNet counterparts, but they are much slower.
This slowdown is caused by the inefficiency of the bypass connection structure of the ShuffleNet unit.
On low-power devices, the bypass connection structure leads to frequent memory and cache switch, which slows down the actual inference speed.
On the contrary, the simple stacking architecture allows FD-MobileNet to leverage memory and cache more efficiently, which contributes to a faster actual inference speed.
These results indicate that FD-MobileNet is effective in actual mobile or embedded applications.

\section{Conclusion}

In this work, we present Fast-Downsampling MobileNet (FD-MobileNet), a highly efficient and accurate network for very limited computational budgets.
FD-MobileNet is built by adopting a fast downsampling strategy in the state-of-the-art MobileNet framework.
Compare with the original MobileNet, the utilization of the fast downsampling scheme allows more channels, which increases the information capacity of the network and contributes to significant performance improvements.
Experiments on the ILSVRC 2012 classification dataset and the PASCAL VOC 2007 detection dataset show that FD-MobileNet consistently outperforms MobileNet under different computational budgets.
Evaluations of the actual inference time demonstrate that FD-MobileNet achieves significant speedup over ShuffleNet on an ARM-based device under the same complexity.
For future work, we plan to adopt the fast downsampling strategy in other compact networks such as ShuffleNet for better performance.

\section{Acknowledgment}

This work is supported by the National Key Research and Development Program of China (2016YFB1000100).

\bibliographystyle{IEEEbib}
\bibliography{main}

\end{document}